# A Conceptual Reasoning Approach to Textual Ellipsis

Udo Hahn, Katja Markert and Michael Strube[1]

**Abstract.** We present a hybrid text understanding methodology for the resolution of textual ellipsis. It integrates conceptual criteria (based on the well-formedness and conceptual strength of role chains in a terminological knowledge base) and functional constraints reflecting the utterances' information structure (based on the distinction between context-bound and unbound discourse elements). The methodological framework for text ellipsis resolution is the centering model that has been adapted to these constraints.

## 1 Introduction

Textual forms of ellipsis and anaphora are a challenging issue for the design of parsers for text understanding systems, since lacking recognition facilities either result in referentially incoherent or invalid text knowledge representations. At the conceptual level, textual ellipsis (also called functional or partial anaphora) relates a quasi-anaphoric expression to its extrasentential antecedent by conceptual attributes (or roles) associated with that antecedent (see, e.g., the relation between *"Ladezeit" (charge time)* and *"Akku" (accumulator)* in (3) and (2)). Thus, it complements the phenomenon of nominal anaphora, where an anaphoric expression is related to its antecedent in terms of conceptual generalization (as, e.g., *"Rechner" (computer)* refers to *"316LT"*, a particular notebook, in (2) and (1)). The resolution of text-level nominal (and pronominal) anaphora contributes to the construction of referentially valid text knowledge bases, while the resolution of textual ellipsis yields referentially coherent text knowledge bases. Both phenomena tend to interact, as evidenced by the example below. *"Akku" (accumulator)* in (2) is a nominal anaphor referring to *"Nickel-Metall-Hydride-Akku" (nickel-metal-hydride accumulator)* in (1), which, when resolved, provides the proper referent for relating *"Ladezeit" (charge time)* in (3) to it.

1. Der *316LT* wird mit einem *Nickel-Metall-Hydride-Akku* bestückt.
   (The *316LT* is – with a *nickel-metal-hydride accumulator* – equipped.)
2. Der *Rechner* wird durch diesen neuartigen *Akku* für 4 Stunden mit Strom versorgt.
   (The *computer* is – because of this new type of *accumulator* – for 4 hours – with power – provided.)
3. Darüberhinaus ist die *Ladezeit* mit 1,5 Stunden sehr kurz.
   (Also – is – the *charge time* of 1.5 hours quite short.)

In the case of textual ellipsis, the missing conceptual link between two discourse elements occurring in adjacent utterances must be inferred in order to establish the local coherence of the discourse (for an early statement of that idea, cf. [3]). In sentence (3), e.g., *"Ladezeit" (charge time)* must be linked up with *"Akku" (accumulator)* from sentence (2). This relation can only be made explicit if conceptual knowledge about the domain, *viz.* the relation *charge-time-of* between the concepts CHARGE-TIME and ACCUMULATOR, is available.

The solution we propose is embedded within the centering model [6], in which textual ellipsis has only been given an insufficient treatment so far. Our approach combines domain and discourse knowledge as well as results from the functional interpretation of the utterances.

On the one hand, language-independent conceptual criteria are based on the well-formedness and conceptual strength of role chains in a terminological knowledge base. On the other hand, we incorporate language-dependent information structure constraints reflecting the context-boundedness or unboundedness of discourse elements within the considered utterances.

## 2 Constraints on Conceptual Linkage

In this section, we will introduce formal and heuristic criteria to determine conceptual links, thus clarifying the notions of *well-formedness* and *strength* of conceptual chains underlying the resolution of textual ellipses. We assume the following conventions to hold in our knowledge base: The concept hierarchy consists of a set of concept names $\mathcal{F}$ = {COMPUTER-SYSTEM, NOTEBOOK, ACCUMULATOR,...} and a subclass relation $isa_\mathcal{F}$ = {(NOTEBOOK, COMPUTER-SYSTEM), (NIMH-ACCUMULATOR, ACCUMULATOR),...} $\subset \mathcal{F} \times \mathcal{F}$. The set of relation names $\mathcal{R}$ = {*has-physical-part, has-accumulator, charge-time-of,...*} contains the labels of all possible conceptual roles. The roles are organized into a hierarchy by the relation $isa_\mathcal{R}$ = {*(has-accumulator, has-physical-part), (charge-time-of, property-of),...*} $\subset \mathcal{R} \times \mathcal{R}$. Throughout the paper, we assume a terminological knowledge representation and reasoning framework (cf. [23] for a survey).

For the identification and evaluation of suitable conceptual links, the ellipsis resolution mechanism is supplied with a *path finder*, which performs an extensive search in the domain knowledge base looking for "well-formed" paths between two concepts, and a *path evaluator*, which selects the "strongests" of the ensuing paths. The path finder applies two basic criteria:

Given two concepts $x, y \in \mathcal{F}$, a series of conceptual relations $r_i \in \mathcal{R}$ ($i = 1,...,n$) and concepts $c_j \in \mathcal{F}$ ($j = 0,...,n$) ($n \in \mathbb{N}$) is admitted as a conceptual path from $x$ to $y$ iff the following *connectivity condition* holds:

- $r_i$ is a (possibly inherited) conceptual role of $c_{i-1}$ with $range(r_i) = c_i$ for all $i = (1,...,n)$;
- $c_0 = x \land (c_n \; isa_\mathcal{F}^* \; y \lor y \; isa_\mathcal{F}^* \; c_n)$, where $isa_\mathcal{F}^*$ denotes the reflexive and transitive closure of $isa_\mathcal{F}$.

Note that no conceptual specialization is allowed at any step of the search except of the end point, thus reducing the complexity of the search. In the following, a *connected conceptual path* like the one above will be denoted by $(r_1 \ldots r_n)$.

Apart from being connective, we require a well-formed path to be *non-cyclic* (cf. Table 1; $r^{-1}$ denotes the inverse of relation $r$).

**Table 1.** Cyclic Path Criterion

| |
|---|
| Cyclic $((r_1 \ldots r_n)) :\Leftrightarrow$ |
| $\exists \; i, j \in \{1, \ldots, n\}: i \neq j \land \exists \; s \in \mathcal{R}: (r_i \; isa_\mathcal{R}^* \; s) \land (r_j \; isa_\mathcal{R}^* \; s^{-1})$ |

This criterion favors a *unidirectional search* in the knowledge base. A cyclic connected conceptual path lacks specificity, as it often only expresses that the end point and the starting point of the search are of a similar, though not necessarily of a semantically related type (for this distinction see, e.g., [18]). For instance, the path (*accumulator-of has-printer*) will be excluded from the search for a path from ACCUMULATOR to PRINTER as *accumulator-of* $isa_\mathcal{R}$ *physical-part-of* and *has-printer* $isa_\mathcal{R}$ *has-physical-part* holds. Thus, the example path carries the information that both, accumulator and printer, are types of hardware, but it does not elucidate any special relationship between these two that an elliptical expression could refer to. A warranted side effect of the exclusion of cyclic patterns is that, as longer paths usually tend to get cyclic, the search terminates without the need to take refuge to *ad hoc* path length restrictions.

Given the set of *well-formed*, i.e., connected and non-cyclic paths, the remaining items of the path list are interpreted by the *path evaluator*. Two criteria are considered in order to select the *strongest* paths among the elements of the path list. One considers the formal inclusion property between well-formed paths, the other introduces semantically plausible path patterns.

**Path Inclusion.** The introduction of a *relative path length* condition is aimed at constraining the overly simplistic counting of nodes in role chains. A well-formed path $p_1 = (r_1 \ldots r_n)$ is *conceptually longer* than another well-formed path $p_2 = (s_1 \ldots s_m)$ iff $p_1$ properly *includes* $p_2$ (see Table 2). The path $p_1$ will then be regarded as being conceptually weaker than $p_2$ and thus be discarded from the path list.

**Table 2.** Path Inclusion Criterion

Includes ($p_1, p_2$) :⇔
$\exists\ i, j \in \{1, \ldots, n\}$: $i \leq j \land (i \neq 1 \lor j \neq n)$
$\land ((r_i \ldots r_j) = (s_1 \ldots s_m))$
$\land ((\text{domain}(r_1)\ isa_\mathcal{F}^*\ \text{domain}(s_1)) \lor (\text{domain}(s_1)\ isa_\mathcal{F}^*\ \text{domain}(r_1)))$
$\land ((\text{range}(r_n)\ isa_\mathcal{F}^*\ \text{range}(s_m)) \lor (\text{range}(s_m)\ isa_\mathcal{F}^*\ \text{range}(r_n)))$

Accordingly, path length considerations cannot be applied to the paths $p_1 = $ (*has-central-unit has-motherboard has-cpu*) and $p_2 = $ (*has-central-unit has-motherboard*) – both being well-formed conceptual paths from NOTEBOOK to PRODUCT. Although $p_2$ is shorter than $p_1$ in the absolute sense (counting role chains or concept nodes), it is not shorter in the relative sense specified above and, thus, not presumed to express a stronger conceptual link (*range (has-cpu)* = CPU and *range (has-motherboard)* = MOTHERBOARD; hence, the last constraint in Table 2 is violated). In contrast, the inclusion criterion is applicable to the paths $p_1 = $ (*has-accumulator price-dm-pair*) and $p_2 = $ (*price-dm-pair*) both leading from NOTEBOOK to PRICE; we regard $p_1$ as being conceptually weaker than $p_2$ given the constraint from Table 2.

**Conceptual Path Patterns.** Finally, we introduce a purely empirical criterion which marks certain paths as being preferred over others in terms of commonsense semantic plausibility. Based on introspective analyses of approximately 60 product reviews from the information technology domain we performed, and evidences reported from several (psycho)linguistic studies (e.g., [1]) , we stipulate certain predefined *path patterns*. From those general path patterns and by virtue of the hierarchical organization of conceptual relations, concrete conceptual role chains can be derived by a simple pattern matching algorithm. These path patterns are used to distinguish between a subset $\mathcal{P}$ of all types of well-formed paths, which is labeled *"plausible"*, another subset $\mathcal{M}$ which is labeled *"metonymic"*, and all remaining paths which are labeled *"implausible"*.

**Plausible Paths.** An important assessment criterion for characterizing relation chains as plausible ones (forming the set $\mathcal{P}$) is that a plausible role chain can always be treated as a single relation. Thus, plausible paths provide a handle for coping with the notorious problem of granularity in knowledge bases. All paths of unit length *1* are included in $\mathcal{P}$, as they are "plausible", by definition (they refer to the conceptual roles directly associated with a concept definition). In addition, we incorporate empirical observations about the transitivity of relations, *part-whole* relations in particular, made by Chaffin [1] and Winston *et al.* [22]. In these studies several subtypes of *part-whole* relations are distinguished, e.g., integral object-component (corresponding to what we call *has-physical-part*), collection-member, mass-portion, process-phase, event-feature and area-place. The claim is made that any of these *sub*relations are transitive, while the most general *part-whole* relation usually is not. In other words, a relation chain containing only relations of one of the above-mentioned subtypes is again *a relation of the same subtype*, whereas a relation chain containing several different types of *part-whole* relations is, in general, not reasonable any more. Following this argument, we have included the path patterns (*has-physical-part**), (*collection-member**), (*mass-portion**), (*process-phase**), (*event-feature**), (*area-place**) and the corresponding inverses like (*physical-part-of**) in $\mathcal{P}$. We will refer to the first six of these patterns as *transitive part-whole patterns*, in short $\mathcal{T}$, and to the inverse patterns as $\mathcal{T}^{-1}$. Apart from the transitive part-whole relations we have included (*spatial containment**) and (*connnection**) in $\mathcal{P}$ (cf. [12]).

**Metonymic Paths.** Following established classifications of metonymies (cf. [14, 5]), we have included the analysis of *whole-for-part*, *part-for-whole*, and *producer-for-product* metonymies in the system. In order to determine path patterns corresponding to these types of metonymies consider the conceptual link between an instance of the concept $C_1$ and an instance of the concept $C_3$, which characterizes a metonymy and thus stands for another instance of a concept $C_2$. A corresponding well-formed conceptual path $p = (r_1 \ldots r_n)$ with $n \in \mathbb{N}, n > 1$, and $r_i \in \mathcal{R}$ ($i = 1,...,n$) must, first, link $C_1$ to $C_2$ via $p_1 = (r_1 \ldots r_{j-1})$ for some $j \in \{2,...,n\}$. $C_2$ is then linked to $C_3$ via $p_2 = (r_j \ldots r_n)$. We restrict the first link $p_1$ to plausible paths in order to provide reasonable metonymic chains only. The second link $p_2$ must express one of the metonymic relations $\mathcal{MS} = \{$*has-part, part-of, produced-by*$\}$, depending on the specific metonymy to be handled[2]. For a *producer-for-product* metonymy, e.g., $j = n$ and $r_n = $ *produced-by* must hold. For a *part-for-whole* or *whole-for-part* metonymy, $j < n$ may be possible as all paths in $\mathcal{T}$ and $\mathcal{T}^{-1}$ (e.g., (*has-physical-part**)) also express a single *has-part* or *part-of* relation (see the explanations of plausible paths above). For notational convenience, we now consider the paths in $\mathcal{T}$ and $\mathcal{T}^{-1}$ as a single relation so that we may write (*has-physical-part**) $isa_\mathcal{R}$ *has-part* or (*event-feature**) $\in \mathcal{MS}$. Thus, we may restrict the above cases of well-formed metonymic paths to the pattern in Table 3. Special path patterns for specific metonymies and metonymic chains can be derived from this general pattern by either instantiating specific metonymic relations or by a recursive application of the predicate.

**Table 3.** Metonymic Path Patterns

Metonymic-Path(($r_1 \ldots r_n$)) :⇔
  ($r_1 \ldots r_n$) $\notin \mathcal{P}$
  $\land ((n > 1 \land (r_1, r_2, \ldots, r_{n-1}) \in \mathcal{P} \land r_n \in \mathcal{MS})$
    $\lor (n > 1 \land (r_2, r_3, \ldots, r_n) \in \mathcal{P} \land r_1 \in \mathcal{MS}^{-1}))$

The markers *"plausible"*, *"metonymic"* and *"implausible"* are finally ranked (cf. Table 4) according to their inherent level of conceptual strength denoted by the relation "$>_{str}$" (conceptually *stronger than*).

---

[2] If the direction of search is reversed (searching from $C_3$ to $C_1$) the corresponding inverse relations must be considered. We refer to these inverse relations as $\mathcal{MS}^{-1} = \{$*part-of, has-part, produces*$\}$. This list of metonymic relations is, of course, incomplete and can be augmented on demand.

**Table 4.** Path Markers Ordered by Conceptual Strength

| "plausible" $>_{str}$ "metonymic" $>_{str}$ "implausible" |
|---|

As a consequence of this ordering, metonymic paths will be excluded from a path list iff plausible paths already exist, while implausible paths will be excluded iff plausible or metonymic paths already exist. At the end of this selection process, only paths of the strongest type are retained in the path list.

To evaluate our approach we selected 80 concept pairs at random from the underlying knowledge base (composed of 459 concepts and 334 relations). We submitted these pairs to the path finder/evaluator and compared the automatically generated conceptual paths with introspective judgments about the kinds of relations linking each pair. The overall error rate was below 5%. The average number of connected paths between two concepts (41.8) was further reduced by the non-cyclicity criterion to 10.4 well-formed paths and by the inclusion criterion (see Table 2) to 2.4. The criterion in Table 4 leads to a final reduction to merely 1.8 paths. Hence, the criteria realize the desired discrimination. We plan a broader evaluation of our approach by running the algorithm on larger-sized knowledge bases in order to test its domain-independence and scaling performance.

All paths which meet the above criteria for two concepts, $x$ and $y$, are contained in a list denoted by $CP_{x,y}$. As, in the case of textual ellipsis, we have to deal with paths leading from the conceptual referent of the elliptical expression to the conceptual referents of several possible antecedents, we usually have to compare pairs of path lists $CP_{x,y}$ and $CP_{x,z}$, where x, y, z $\in \mathcal{F}$ ($y \neq z$). Fortunately, the same criteria can be applied to path lists as those we used for evaluating paths linking single concepts. As all paths in $CP_{x,y}$ and $CP_{x,z}$ were computed by the path finder, they already fulfill the connectivity and non-cyclicity condition. The inclusion criterion (see Table 2) cannot be applied to any path $p_1 \in CP_{x,y}$ and $p_2 \in CP_{x,z}$, as $p_1$ and $p_2$ have different end points, by definition. However, the criterion which ranks conceptual paths according to their associated path markers is applicable, as all paths in a single CP list have the same marker. A function, *PathMarker*($CP_{i,j}$), yields as its value either *"plausible"*, *"metonymic"* or *"implausible"* depending on the type of paths it contains. We may now apply the same ordering of path markers as in Table 4 in order to compare two CP lists (cf. Table 5).

**Table 5.** Path Lists Compared by Conceptual Strength

| isStrongerThan ($CP_{x,y}$, $CP_{x,z}$) :⇔ <br>     PathMarker($CP_{x,y}$) $>_{str}$ PathMarker($CP_{x,z}$) <br><br> equallyStrongAs ($CP_{x,y}$, $CP_{x,z}$) :⇔ <br>     PathMarker($CP_{x,y}$) = PathMarker($CP_{x,z}$) |
|---|

## 3 Functional Constraints on Centers

Conceptual criteria are of tremendous importance, but they are not sufficient for proper ellipsis resolution. Additional criteria have to be supplied in the case of equal strength of CP lists for alternative antecedents. We therefore incorporate into our model criteria which relate to the functional information structure of utterances using the methodological framework of the well-known *centering* model [6].

The theory of centering is intended to model the local coherence of discourse, i.e., coherence among the utterances in a particular discourse segment (say, a paragraph of a text). Each utterance $U_i$ in a discourse segment is assigned a set of *forward-looking centers*, $C_f(U_i)$, and a unique *backward-looking center*, $C_b(U_i)$. The elements of $C_f(U_i)$ are partially ordered to reflect relative prominence in $U_i$. The most highly ranked element of $C_f(U_i)$ that is *realized* in $U_{i+1}$ (i.e., is associated with an expression that has a valid semantic interpretation) is the $C_b(U_{i+1})$. The ranking imposed on the elements of the $C_f$ reflects the assumption that the most highly ranked element of $C_f(U_i)$ is the most preferred antecedent of an anaphoric expression in $U_{i+1}$, while the remaining elements are partially ordered according to decreasing preference for establishing referential links.

The theory of centering, in addition, defines several transition relations across pairs of adjacent utterances (e.g., continuation, retention, smooth and rough shift), which differ from each other according to the degree by which successive backward-looking centers are confirmed or rejected, and, if they are confirmed, whether they correspond to the most highly ranked element of the current forward-looking centers or not. The theory claims that to the extent a discourse adheres to all these centering constraints (e.g., realization constraints on pronouns, preferences among types of center transitions), its local coherence will increase and the inference load placed upon the hearer will decrease. Therefore, the tremendous importance of fleshing out the relevant and most restrictive, though still general centering constraints.

The main difference between Grosz *et al.*'s seminal work [6] and our proposal (see [20]) concerns the criteria for ranking the forward-looking centers. While Grosz *et al.* assume that *grammatical roles* are the major determinant for the ranking on the $C_f$, we claim that for languages with relatively free word order (such as German), it is the *functional information structure (IS)* of the utterance in terms of the context-boundedness or unboundedness of its discourse elements. The centering data structures and the notion of context-boundedness can be used to redefine Danes' [4] trichotomy between *given information*, *theme* and *new information* (which he considers equivalent to *rheme*). The $C_b(U_n)$, the most highly ranked element of $C_f(U_{n-1})$ realized in $U_n$, corresponds to the element which represents the *given* information. The *theme* of $U_n$ is represented by the preferred center $C_p(U_n)$, the most highly ranked element of $C_f(U_n)$. The *theme/rheme hierarchy* of $U_n$ is determined by the $C_f(U_{n-1})$: the rhematic elements of $U_n$ are the ones not contained in $C_f(U_{n-1})$ (unbound discourse elements) – they express the *new information* in $U_n$. The ones contained in $C_f(U_{n-1})$ and $C_f(U_n)$ (bound discourse elements) are thematic, with the theme/rheme hierarchy corresponding to the ranking in the $C_f$s.

## 4 Grammatical Predicates for Textual Ellipsis

The grammar formalism we use (cf. [8] for a survey) is based on dependency relations between lexical heads and modifiers. The dependency specifications allow a tight integration of linguistic (grammar) and conceptual knowledge (domain model), thus making powerful terminological reasoning facilities directly available for the parsing process[3]. The resolution of textual ellipses is based on two major criteria, a conceptual and a structural one. The conceptual strength criterion for role chains is already specified in Table 5. The structural condition is embodied in the predicate *isPotentialEllipticAntecedent* (cf. Table 6). The elliptical phrase which occurs in the *n*-th utterance is restricted to be a definite NP and the antecedent must be one of the forward-looking centers of the preceding utterance (note that $C_f$s contain only conceptual referents of nouns and pronouns).

The predicate *PreferredConceptualBridge* (cf. Table 7) combines both criteria. A lexical item *y* is determined as the proper antecedent

---

[3] We assume the following conventions to hold: $\mathcal{C}$ = {Word, Nominal, Noun, PronPersonal,...} denotes the set of word classes, and $isa_\mathcal{C}$ = {(Nominal, Word), (Noun, Nominal), (PronPersonal, Nominal),...} $\subset \mathcal{C} \times \mathcal{C}$ denotes the subclass relation which yields a hierarchical ordering among these classes. Furthermore, *object.r* refers to the instance in the text knowledge base denoted by the linguistic item *object* and *object.c* refers to the corresponding concept class C. *Head* denotes a structural relation within dependency trees, viz. *x* being the head of modifier *y*.

**Table 6.** Potential Elliptical Antecedent

isPotentialEllipticAntecedent (y, x, n) :⇔
y $isa_C$* Nominal ∧ x $isa_C$* Noun
∧ ∃ z: (x $head$ z ∧ z $isa_C$* DetDefinite)
∧ x ∈ $U_n$ ∧ y.r ∈ $C_f(U_{n-1})$

of the elliptic expression *x* iff it is a potential antecedent and if there exists no alternative antecedent *z* whose conceptual strength relative to *x* exceeds that of *y* or, if their conceptual strength is equal, whose strength of preference under the *IS* relation is higher than that of *y*. "$>_{IS}$" defines (cf. [20] for an in-depth treatment) a strict order on the conceptual/semantic items of $C_f$ reflecting the functional information structure of the utterance $U_n$ in which their linguistic counterparts, *viz. z* and *y*, occur.

**Table 7.** Preferred Conceptual Bridge for an Elliptical Expression

PreferredConceptualBridge (y, x, n) :⇔
isPotentialEllipticAntecedent (y, x, n)
∧ ¬∃ z : isPotentialEllipticAntecedent (z, x, n)
∧ (isStrongerThan (CP$_{x.c,z.c}$, CP$_{x.c,y.c}$)
∨ (equallyStrongAs (CP$_{x.c,z.c}$, CP$_{x.c,y.c}$) ∧ z $>_{IS}$ y ) )

## 5 Text Ellipsis Resolution

The resolution of textual ellipses depends on the results of the foregoing resolution of nominal anaphors [19] and the termination of the semantic interpretation of the current utterance. It will only be triggered at the occurrence of the definite noun phrase *NP* when *NP* is not a nominal anaphor and (the conceptual referent of the) *NP* is only connected via certain types of relations (e.g., *has-property, has-physical-part*)[4] to referents denoted in the current utterance at the conceptual level.

We will illustrate our approach to text ellipsis resolution, referring to the already introduced text fragment (1) – (3). (3) contains the definite noun phrase *"die Ladezeit"*. At the conceptual level, *"Ladezeit" (charge time)* does not subsume any element of the forward-looking centers of the previous utterance ($C_f(U_2)$ = [316LT, ACCUMULATOR, TIME-UNIT-PAIR, POWER]). Thus, the anaphora test fails; the conceptual referent of *"die Ladezeit"* has also not been integrated in terms of a significant relation into the conceptual representation of the utterance as a result of its semantic interpretation. Consequently, the search for an antecedent of the textual ellipsis is triggered.

The forward-looking centers of the previous sentence are tested for the predicate *PreferredConceptualBridge*. In this case, the instance 316LT (the conceptual referent of the nominal anaphor *"der Rechner" (the computer)*, which has already been properly resolved) is related to CHARGE-TIME (the concept denoting *"Ladezeit"*) via a *metonymic path*, viz. *(charge-time-of accumulator-of)*. This path corresponds to a *whole-for-part* metonymy, as *charge time* is a direct property of an accumulator and therefore only a mediated property of a computer as a whole. In contrast, the concept ACCUMULATOR is related to CHARGE-TIME via a *plausible path* (viz. *charge-time-of*). As plausible paths are the strongest type of conceptual paths, none of the items following in the centering list can be preferred as the antecedent of *"Ladezeit" (charge time)* over *"Akku" (accumulator)* (cf. the constraint from Table 7). Hence, the remaining concepts in the $C_f$ list (*viz.* TIME-UNIT-PAIR and POWER) need no longer be considered as potential antecedents. An appropriate update links the corresponding instances via the role *charge-time-of* and, thus, local coherence is established at the conceptual level of the text knowledge base. A fully worked out parsing example together with a discussion of a medium-sized performance evaluation of the criteria considered for ellipsis resolution is given in [9].

---

[4] The distinction between roles and their inverses becomes crucial for already established relations like *has-property* (subsuming *charge-time*, etc.) or *has-physical-part* (subsuming *has-accumulator*, etc.). The instantiation of these relations does not block the triggering of the resolution procedure for textual ellipsis (e.g., ACCUMULATOR – *charge-time* – CHARGE-TIME), whereas instantiations of their inverses, we here refer to as *POF-type relations*, e.g., *property-of* (subsuming *charge-time-of*, etc.) or *physical-part-of* (subsuming *accumulator-of*, etc.), do (e.g., CHARGE-TIME – *charge-time-of* – ACCUMULATOR). This is simply due to the fact that the semantic interpretation of a phrase like *"the charge time of the accumulator"* already leads to the creation of the *POF*-type relation the resolution mechanism for textual ellipsis is supposed to determine. This is opposed to the interpretation of its elliptified counterpart *"the charge time"* in sentence (3), where the genitive object *"[of the accumulator]"* is elided and, thus, the role *charge-time-of* remains uninstantiated.

## 6 Comparison with Related Approaches

Searching links in a taxonomic hierarchy is a problem which has often been tackled by spreading activation or marker passing approaches. The paradigm of path finding and evaluating they propose has obvious parallels to our approach. The criteria used in spreading activation models for finding and evaluating paths, however, are mostly based on numerical restrictions, e.g., on weights [2] or path lengths [10]. This is problematic, as the foundation and derivation of these numbers is usually not made explicit.

The abduction-based approach to inferencing underlying the TACITUS system [11] also refers to weights and costs and, thus, shares some similarity with marker passing proposals [11, p. 122]. The crucial problem, however, still unsolved in this logically very principled framework concerns a proper choice methodology for fixing appropriate costs for specific assumptions on which, among other factors, textual ellipsis resolution is primarily based.

A pattern-based approach to inferencing (including textual ellipsis resolution) has also been put forward by Norvig [16]. Unlike Norvig's proposal to define path patterns solely in terms of "formal" link criteria in a knowledge base whose patterns are simply matched against the links being passed, our definitions of path patterns take the semantic hierarchy of relations and their compositional properties (like transitivity) into account. This allows for a semantically motivated preference ranking of the paths by treating the phenomena of granularity (corresponding to plausible paths) and metonymy (corresponding to metonymic paths) in a unified search algorithm. Although Norvig makes a strong point concerning his use of path patterns (instead of marker energy) to guide the search in a knowledge base, the definitions of path patterns he gives are not restrictive enough. Additional numerical rules for coping with combinatorial search problems (e.g., an antipromiscuity rule) still have to be supplied, whereas our path patterns are complemented by structural formal criteria which do not rely upon numerical restrictions in any way. The cyclicity criterion, e.g., leads to path length (and thus granularity) independence and the inclusion criterion further abstracts from node counting.

As far as text-level processing is concerned, the framework of DRT [13], at first sight, constitutes a particularly strong alternative to our approach. The machinery of DRT, however, might work well for (pro)nominal anaphora, but faces problems when elliptical text phenomena are to be interpreted (though [21] has recently made an attempt to deal with restricted forms of textual ellipsis in the DRT context). This shortcoming is simply due to the fact that DRT is basically a semantic theory, not a full-fledged model for text understanding. In particular, it lacks any systematic connection to well-developed reasoning systems accounting for conceptual domain knowledge. Actually, the sort of constraints we consider seem much more rooted in encyclopedic knowledge than are they of a primarily semantic nature anyway.

Unlike DRT, the centering model [6] to which we adhere is not considered a semantic theory but rather a discourse processing model which lends itself to an easy integration into a text understanding framework. Still, it does not provide for well-developed methods for textual ellipsis resolution. Grosz et al. rather sketchily point to the difference between the relations *directly realizes* and *realizes* whose precise definition they suggest depends on the semantic theory one adopts [6, p.209]. We have shown, however, that there are a lot of constraints at the conceptual level which cannot reasonably be accounted for by semantic theories.

Only few NLP systems exist which deal with textual ellipsis in a dedicated way. For example, the PUNDIT system [17] provides a fairly restricted solution in that only direct conceptual links between the concept denoted by the antecedent and the elliptical expression are considered ("plausible" paths of length *1*, in our terminology). The approach reported in this paper also extends our own previous work on textual ellipses [7] by the incorporation of an elaborated model of functional preferences on $C_f$ elements which constrains the set of possible antecedents according to information structure criteria.

## 7 Conclusions

In this paper, we have outlined a model of textual ellipsis resolution. It considers conceptual criteria to be of primary importance and provides conceptual well-formedness and strength criteria for role chains in a terminological knowledge base in order to assess the plausibility of various possible antecedents as proper bridges [3] to elliptical expressions. Functional constraints based on the utterances' information structure contribute further restrictions on proper elliptical antecedents and require a basic revision of the centering model.

The two principled difficulties inherent in every network-based symbolic knowledge representation approach are its dependency on hand-crafted and often domain-specific knowledge and the exorbitant costs for unconstrained search (often limiting the scalability of the approach). We cope with the first of these problems by postulating two formal criteria, *viz.* the non-cyclicity and the inclusion condition, which only depend on features present in almost any knowledge representation language, i.e., *isa* links, domains and ranges of relations and inverse relations. No reference to a *specific* network structure or a *specific* domain is made. The definition of patterns relies mainly on structural properties of semantic relations, which are entirely domain-independent. However, this heuristic criterion is only effective when the underlying knowledge base is built on top of a clear taxonomy of relations (although this taxonomy and the corresponding path patterns can be specified in ways differing from ours). The expensiveness of the search has already been reduced significantly by testing the non-cyclicity of the paths during the search. We are currently experimenting with an additional search constraint, which limits the search to certain *dynamically narrowed regions* in a knowledge base, likely to make the algorithm efficiently executable even on larger knowledge bases.

The ellipsis handler has been implemented in Smalltalk as part of a comprehensive text parser for German, which is interfaced to the LOOM system [15]. Besides the information technology domain (this knowledge base currently contains approximately 800 concept/role specifications), experiments with our parser have also been successfully run on medical domain texts (the corresponding medical domain knowledge base currently contains approximately 500 concept/role specifications). These results indicate that the heuristics we have been developing are not bound to a particular domain.

**Acknowledgments.** We would like to thank our colleagues in the $\mathcal{CLIF}$ group for fruitful discussions. M. Strube's contribution has been funded by *LGFG Baden-Württemberg*, while K. Markert is supported by a grant from DFG within the Freiburg University Graduate Program on *"Human and Artificial Intelligence"*. We also gratefully acknowledge the provision of the LOOM system from USC/ISI.

## REFERENCES

[1] Roger Chaffin, `The concept of a semantic relation', in *Frames, Fields and Contrasts*, eds., A. Lehrer and E.F. Kittay, 253–288, Hillsdale, N.J.: Lawrence Erlbaum, (1992).

[2] Eugene Charniak, `A neat theory of marker passing', in *Proc. of AAAI-86*, volume 1, pp. 584–588, (1986).

[3] Herbert H. Clark, `Bridging', in *Proc. of the Conference on Theoretical Issues in Natural Language Processing, Cambridge, MA*, pp. 169–174, (1975).

[4] Frantisek Danes, `Functional sentence perspective and the organization of the text', in *Papers on Functional Sentence Perspective*, ed., F. Danes, 106–128, Prague: Academia, (1974).

[5] Dan C. Fass, `met*: A method for discriminating metonymy and metaphor by computer', *Computational Linguistics*, **17**(1), 49–90, (1991).

[6] Barbara J. Grosz, Aravind K. Joshi, and Scott Weinstein, `Centering: A framework for modeling the local coherence of discourse', *Computational Linguistics*, **21**(2), 203–225, (1995).

[7] Udo Hahn, `Making understanders out of parsers: Semantically driven parsing as a key concept for realistic text understanding applications', *International Journal of Intelligent Systems*, **4**(3), 345–393, (1989).

[8] Udo Hahn, Susanne Schacht, and Norbert Bröker, `Concurrent, object-oriented dependency parsing: The *ParseTalk* model.', *International Journal of Human-Computer Studies*, **41**(1/2), 179–222, (1994).

[9] Udo Hahn, Michael Strube, and Katja Markert, `Bridging textual ellipses', in *Proc. of COLING-96*, (1996).

[10] Graeme Hirst, *Semantic Interpretation and the Resolution of Ambiguity*, Cambridge, UK: Cambridge University Press, 1987.

[11] Jerry R. Hobbs, Mark E. Stickel, Douglas E. Appelt, and Paul Martin, `Interpretation as abduction', *Artificial Intelligence*, **63**, 69–142, (1993).

[12] Michael N. Huhns and Larry M. Stephens, `Plausible inferencing using extended composition', in *Proc. of IJCAI-89*, volume 2, pp. 1420–1425, (1989).

[13] Hans Kamp and Uwe Reyle, *From Discourse to Logic*, Dordrecht: Kluwer, 1993.

[14] George Lakoff, *Women, Fire, and Dangerous Things. What Categories Reveal about the Mind*, Chicago University Press, Chicago, IL, 1987.

[15] Robert MacGregor and Raymond Bates. The LOOM Knowledge Representation Language. (ISI/RS-87-18) USC/ISI, 1987.

[16] Peter Norvig, `Marker passing as a weak method for inferencing', *Cognitive Science*, **13**(4), 569–620, (1989).

[17] Martha S. Palmer, Deborah A. Dahl, Rebecca J. Schiffman, and Lynette Hirschman, `Recovering implicit information', in *Proc. of ACL-86*, pp. 10–19, (1986).

[18] Phil Resnik, `Using information content to evaluate semantic similarity in a taxonomy', in *Proc. of IJCAI-95*, volume 1, pp. 448–453, (1995).

[19] Michael Strube and Udo Hahn, `*ParseTalk* about sentence- and text-level anaphora', in *Proc. of EACL-95*, pp. 237–244, (1995).

[20] Michael Strube and Udo Hahn, `Functional centering', in *Proc. of ACL-96*, (1996).

[21] Hajime Wada, `A treatment of functional definite descriptions', in *Proc. of COLING-94*, volume 2, pp. 789–795, (1994).

[22] Morton Winston, Roger Chaffin, and Douglas Herrmann, `A taxonomy of part-whole-relations', *Cognitive Science*, **11**, 417–444, (1987).

[23] William A. Woods and James G. Schmolze, `The KL-ONE family', *Computers & Mathematics with Applications*, **23**(2-5), 133–177, (1992).